\documentclass[letterpaper]{sig-alternate-2013}
\usepackage[
  pass,
]{geometry}

\permission{\copyright 2017 International World Wide Web Conference Committee \\ (IW3C2), published under Creative Commons CC BY 4.0 License.}
\conferenceinfo{WWW 2017,}{April 3--7, 2017, Perth, Australia.}
\copyrightetc{ACM \the\acmcopyr}
\crdata{978-1-4503-4913-0/17/04. \\
http://dx.doi.org/10.1145/3038912.3052642 \\
\includegraphics{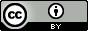}
}

\begin{document}






%
\conferenceinfo{WWW}{'17 Perth, Australia}

\title{Information Extraction in Illicit Web Domains}

%
%
%
%
%

\numberofauthors{2} 
%
\author{
%
%
\alignauthor
Mayank Kejriwal\\
       \affaddr{Information Sciences Institute}\\
       \affaddr{USC Viterbi School of Engineering}\\
       \email{kejriwal@isi.edu}
\alignauthor
Pedro Szekely\\
       \affaddr{Information Sciences Institute}\\
       \affaddr{USC Viterbi School of Engineering}\\
       \email{pszekely@isi.edu}
}


\maketitle
\begin{abstract}
Extracting useful entities and attribute values from illicit domains such as human trafficking is a challenging problem with the potential for widespread social impact. Such domains employ atypical language models, have `long tails' and suffer from the problem of concept drift. In this paper, we propose a lightweight, feature-agnostic Information Extraction (IE) paradigm specifically designed for such domains. Our approach uses raw, unlabeled text from an initial corpus, and a few (12-120) seed annotations per domain-specific attribute, to learn robust IE models for unobserved pages and websites. Empirically, we demonstrate that our approach can outperform feature-centric Conditional Random Field baselines by over 18\% F-Measure on five annotated sets of real-world human trafficking datasets in both low-supervision and high-supervision settings. We also show that our approach is demonstrably robust to concept drift, and can be efficiently bootstrapped even in a serial computing environment. 
\end{abstract}

%
%


%
%

%
%



\section{Introduction}\label{introduction}
Building knowledge graphs (KG) over Web corpora is an important problem that has galvanized effort from multiple communities over two decades \cite{kgc}, \cite{kg1}. Automated knowledge graph construction from Web resources involves several different phases. The first phase involves \emph{domain discovery}, which constitutes identification of sources, followed by crawling and scraping of those sources \cite{domaindiscovery1}. A contemporaneous \emph{ontology engineering} phase is the identification and design of key classes and properties in the domain of interest (the \emph{domain ontology}) \cite{ontologyengineering}.

Once a set of (typically unstructured) data sources has been identified, an \emph{Information Extraction} (IE) system needs to extract structured data from each page in the corpus \cite{IE}, \cite{webIE}, \cite{wrapperIE}, \cite{stanfordner}. In IE systems based on statistical learning, \emph{sequence labeling} models like Conditional Random Fields (CRFs) can be trained and used for tagging the scraped text from each data source with terms from the domain ontology \cite{crf1}, \cite{stanfordner}. With enough data and computational power, deep neural networks can also be used for a range of collective natural language tasks, including chunking and extraction of named entities and relationships \cite{multitask}.

While IE has been well-studied both for cross-domain Web sources (e.g. Wikipedia) and for traditional domains like biomedicine \cite{wikipediaIE}, \cite{biomedicalNER}, it is less well-studied (Section \ref{relatedwork}) for \emph{dynamic} domains that undergo frequent changes in content and structure. Such domains include news feeds, social media, advertising, and online marketplaces, but also \emph{illicit} domains like human trafficking. Automatically constructing knowledge graphs containing important information like ages (of human trafficking victims), locations, prices of services and posting dates over such domains could have widespread social impact, since law enforcement and federal agencies could query such graphs to glean rapid insights \cite{dig}.

Illicit domains pose some formidable challenges for traditional IE systems, including deliberate information \emph{obfuscation}, non-random misspellings of common words, high occurrences of out-of-vocabulary and uncommon words, frequent (and non-random) use of Unicode characters, sparse content and heterogeneous website structure, to only name a few \cite{dig}, \cite{alvari}, \cite{ben}. While some of these characteristics are shared by more traditional domains like chat logs and Twitter, both information obfuscation and extreme content heterogeneity are unique to illicit domains. While this paper only considers the human trafficking domain, similar kinds of problems are prevalent in other illicit domains that have a sizable Web (including Dark Web) footprint, including terrorist activity, and sales of illegal weapons and counterfeit goods \cite{darkweb}. 

As real-world illustrative examples, consider the text fragments \emph{`Hey gentleman im neWYOrk and i'm looking for generous...'} and \emph{`AVAILABLE NOW! ?? - (4 two 4) six 5 two - 0 9 three 1 - 21'}. In the first instance, the correct extraction for a \emph{Name} attribute is \emph{neWYOrk}, while in the second instance, the correct extraction for an \emph{Age} attribute is \emph{21}. It is not obvious what features should be engineered in a statistical learning-based IE system to achieve robust performance on such text.  

To compound the problem, \emph{wrapper induction} systems from the Web IE literature cannot always be applied in such domains, as many important attributes can only be found in text descriptions, rather than template-based Web extractors that wrappers traditionally rely on \cite{wrapperIE}. Constructing an IE system that is robust to these problems is an important first step in delivering structured knowledge bases to investigators and domain experts. 
\begin{figure*}
\centering
\includegraphics[height=2.8in, width=6.7in]{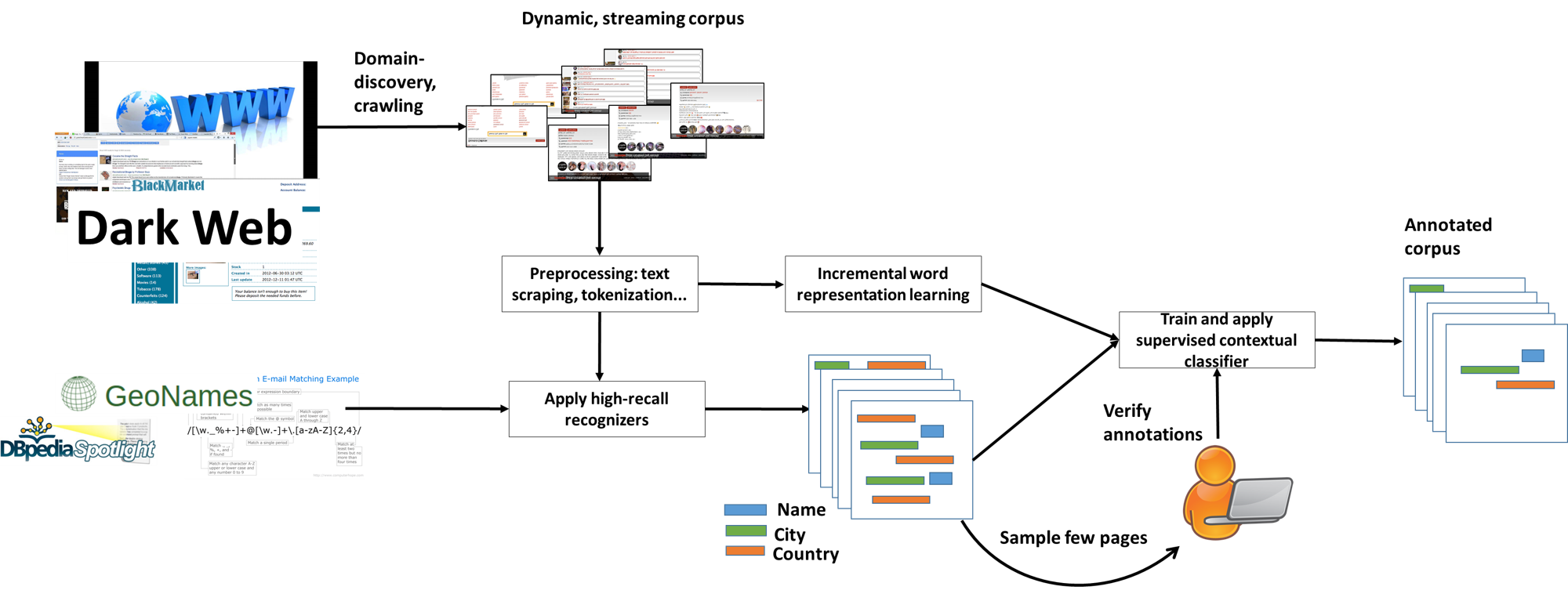}
\caption{A high-level overview of the proposed information extraction approach}\label{fig:approach}
\end{figure*}

In this paper, we study the problem of robust information extraction in dynamic, illicit domains with unstructured content that does not necessarily correspond to a typical natural language model, and that can vary tremendously between different Web domains, a problem denoted more generally as \emph{concept drift} \cite{conceptdrift}. Illicit domains like human trafficking also tend to exhibit a `long tail'; hence, a comprehensive solution should not rely on information extractors being tailored to pages from a small set of Web domains. 

There are two main technical challenges that such domains present to IE systems. First, as the brief examples above illustrate, feature engineering in such domains is difficult, mainly due to the atypical (and varying) representation of information. Second, investigators and domain experts require a \emph{lightweight} system that can be quickly bootstrapped. Such a system must be able to generalize from few ($\approx$10-150) manual annotations, but be incremental from an engineering perspective, especially since a given illicit Web page can quickly (i.e. within hours) become obsolete in the real world, and the search for leads and information is always ongoing. In effect, the system should be designed for streaming data. 

We propose an information extraction approach that is able to address the challenges above, especially the variance between Web pages and the small training set per attribute, by combining two sequential techniques in a novel paradigm. The overall approach is illustrated in Figure \ref{fig:approach}. First, a \emph{high-recall recognizer}, which could range from an exhaustive Linked Data source like GeoNames (e.g. for extracting locations) to a simple regular expression (e.g. for extracting ages), is applied to each page in the corpus to derive a set of \emph{candidate annotations} for an attribute per page. In the second step, we train and apply a supervised feature-agnostic classification algorithm, based on learning word representations from random projections, to classify each candidate as correct/incorrect for its attribute. 

{\bf Contributions} We summarize our main contributions as follows: (1) We present a lightweight feature-agnostic information extraction system for a highly heterogeneous, illicit domain like human trafficking. Our approach is simple to implement, does not require extensive parameter tuning, infrastructure setup and is incremental with respect to the data, which makes it suitable for deployment in streaming-corpus settings. (2) We show that the approach shows good generalization even when only a small corpus is available after the initial domain-discovery phase, and is robust to the problem of concept drift encountered in large Web corpora. (3) We test our approach extensively on a real-world human trafficking corpus containing hundreds of thousands of Web pages and millions of unique words, many of which are rare and highly domain-specific. Evaluations show that our approach outperforms traditional Named Entity Recognition baselines that require manual feature engineering. Specific empirical highlights are provided below.

{\bf Empirical highlights} Comparisons against CRF baselines based on the latest Stanford Named Entity Resolution system (including pre-trained models as well as new models that we trained on human trafficking data) show that, on average, across five ground-truth datasets, our approach outperforms the next best system on the recall metric by about 6\%, and on the F1-measure metric by almost 20\% in low-supervision settings (30\% training data), and almost 20\% on both metrics in high-supervision settings (70\% training data). Concerning efficiency, in a serial environment, we are able to derive word representations on a 43 million word corpus in under an hour. Degradation in average F1-Measure score achieved by the system is less than 2\% even when the underlying raw corpus expands by a factor of 18, showing that the approach is reasonably robust to concept drift.     

{\bf Structure of the paper} Section \ref{relatedwork} describes some related work on Information Extraction. Section \ref{approach} provides details of key modules in our approach. Section \ref{evaluations} describes experimental evaluations, and Section \ref{conclusion} concludes the work.

\section{Related Work}\label{relatedwork}
Information Extraction (IE) is a well-studied research area both in the Natural Language Processing community and in the World Wide Web, with the reader referred to the survey by Chang et al. for an accessible coverage of Web IE approaches \cite{IEsurvey}. In the NLP literature, IE problems have predominantly been studied as Named Entity Recognition and Relationship Extraction \cite{stanfordner}, \cite{relex}. The scope of Web IE has been broad in recent years, extending from wrappers to Open Information Extraction (OpenIE) \cite{wrapperIE}, \cite{openIE}. 

In the Semantic Web, domain-specific extraction of  entities and properties is a fundamental aspect in constructing instance-rich knowledge bases (from unstructured corpora) that contribute to the Semantic Web vision and to ecosystems like Linked Open Data \cite{lod}, \cite{SWvision}. A good example of such a system is Lodifier \cite{lodifier}. This work is along the same lines, in that we are interested in user-specified attributes and wish to construct a knowledge base (KB) with those attribute values using raw Web corpora. However, we are not aware of any IE work in the Semantic Web that has used word representations to accomplish this task, or that has otherwise outperformed state-of-the-art systems without manual feature engineering.  

The work presented in this paper is structurally similar to the geolocation prediction system (from Twitter) by Han et al. and also ADRMine, an adverse drug reaction (ADR) extraction system from social media  \cite{rev1ref1}, \cite{rev1ref2}. Unlike these works, our system is not optimized for specific attributes like locations and drug reactions, but generalizes to a range of attributes. Also, as mentioned earlier, illicit domains involve challenges not characteristic of social media, notably information obfuscation. 

In recent years, state-of-the-art results have been achieved in a variety of NLP tasks using word representation methods like neural embeddings \cite{word2vec}. Unlike the problem covered in this paper, those papers typically assume an existing KB (e.g. Freebase), and attempt to infer additional facts in the KB using word representations. In contrast, we study the problem of constructing and populating a KB per domain-specific attribute \emph{from scratch} with only a small set of initial annotations from crawled Web corpora.  

The problem studied in this paper also has certain resemblances to OpenIE \cite{openIE}. One assumption in OpenIE systems is that a given fact (codified, for example, as an RDF triple) is observed in multiple pages and contexts, which allows the system to learn new `extraction patterns' and rank facts by confidence. In illicit domains, a `fact' may only be observed once; furthermore, the arcane and high-variance language models employed in the domain makes direct application of any extraction pattern-based approach problematic. 
To the best of our knowledge, the specific problem of devising feature-agnostic, low-supervision IE approaches for illicit Web domains has not been studied in prior work. 

\section{Approach}\label{approach}

Figure \ref{fig:approach} illustrates the architecture of our approach. The input is a Web corpus containing relevant pages from the domain of interest, and \emph{high-recall recognizers} (described in Section \ref{recognizers}) typically adapted from freely available Web resources like Github and GeoNames. In keeping with the goals of this work, we do not assume that this initial corpus is static. That is, following an initial short set-up phase, more pages are expected to be added to the corpus in a streaming fashion. Given a set of pre-defined attributes (e.g. City, Name, Age) and  around 10-100 manually verified annotations for each attribute, the goal is to learn an IE model that accurately extracts attribute values from each page in the corpus without relying on expert feature engineering. Importantly, while the pages are single-\emph{domain} (e.g. human trafficking) they are \emph{multi-Web domain}, meaning that the system must not only handle pages from new websites as they are added to the corpus, but also \emph{concept drift} in the new pages compared to the initial corpus.  

\subsection{Preprocessing}\label{preprocessing}

The first module in Figure \ref{fig:approach} is an automated pre-processing algorithm that takes as input a streaming set of HTML pages. In real-world illicit domains, the key information of interest to investigators (e.g. names and ages) typically occurs either in the \emph{text} or the \emph{title} of the page, not the template of the website. Even when the information occasionally occurs in a template, it must be appropriately disambiguated to be useful\footnote{For example, `Virginia' in South Africa vs. `Virginia' in the US.}. Wrapper-based IE systems \cite{wrapperIE} are often inapplicable as a result. As a first step in building a more suitable IE model, we scrape the text from each HTML website by using a publicly available text extractor called the \emph{Readability Text Extractor}\footnote{\url{https://www.readability.com/developers/api}} (RTE). Although multiple tools\footnote{An informal comparison may be accessed at \url{https://www.diffbot.com/benefits/comparison/}} are available for text extraction from HTML \cite{te1}, our early trials showed that RTE is particularly suitable for noisy Web domains, owing to its tuneability, robustness and support for developers. We tune RTE to achieve \emph{high recall}, thus ensuring that the relevant text in the page is captured in the scraped text with high probability. Note that, because of the varied structure of websites, such a setting also introduces noise in the scraped text (e.g. wayward HTML tags). Furthermore, unlike natural language documents, scraped text can contain many irrelevant numbers, Unicode and punctuation characters, and may not be regular. Because of the presence of numerous tab and newline markers, there is no obvious natural language \emph{sentence} structure in the scraped text\footnote{We also found sentence ambiguity in the actual text displayed on the browser-rendered website (in a few human trafficking sample pages), due to the language models employed in these pages.}. In the most general case, we found that RTE returned a set of strings, with each string corresponding to a set of sentences.

To serialize the scraped text as a list of tokens, we use the word and sentence tokenizers from the NLTK package on each RTE string output \cite{nltk}. We apply the sentence tokenizer first, and to each sentence returned (which often does not correspond to an actual sentence due to rampant use of extraneous punctuation characters) by the sentence tokenizer, we apply the standard NLTK word tokenizer. The final output of this process is a list of tokens. In the rest of this section, this list of tokens is assumed as representing the HTML page from which the requisite attribute values need to be extracted. 

\subsection{Deriving Word Representations}\label{embeddings}

In principle, given some annotated data, a sequence labeling model like a Conditional Random Field (CRF) can be trained and applied on each block of scraped text to extract values for each attribute \cite{crf1}, \cite{stanfordner}. In practice, as we empirically demonstrate in Section \ref{evaluations}, CRFs prove to be problematic for illicit domains. First, the size of the training data available for each CRF is relatively small, and because of the nature of illicit domains, methods like distant supervision or crowdsourcing cannot be used in an obvious timely manner to elicit annotations from users.
A second problem with CRFs, and other traditional machine learning models, is the careful feature engineering that is required for good performance. With small amounts of training data, good features are essential for generalization. In the case of illicit domains, it is not always clear what features are appropriate for a given attribute. Even common features like capitalization can be misleading, as there are many capitalized words in the text that are not of interest (and vice versa). 

To alleviate feature engineering and manual annotation effort, we leverage the entire raw corpus in our model learning phase, rather than just the pages that have been annotated. Specifically, we use an unsupervised algorithm to represent each word in the corpus in a low-dimensional \emph{vector space}. Several algorithms exist in the literature for deriving such representations, including neural embedding algorithms such as Word2vec \cite{word2vec} and the algorithm by Bollegala et al. \cite{bollegala2015}, as well as simpler alternatives \cite{randomindexing}. 

Given the dynamic nature of streaming illicit-domain data, and the numerous word representation learning algorithms in the literature, we adapted the \emph{random indexing} (RI) algorithm for deriving contextual word representations \cite{randomindexing}. Random indexing methods mathematically rely on the Johnson-Lindenstrauss Lemma, which states that if points in a vector space are of sufficiently high dimension, then they may be projected into a suitable lower-dimensional space in a way which approximately preserves the distances between the points.

The original random indexing algorithm was designed for incremental dimensionality reduction and text mining applications. We adapt this algorithm for learning word representations in illicit domains.  
Before describing these adaptations, we define some key concepts below.   

\newdef{definition}{Definition}
\begin{definition}\label{contextvec}
Given parameters $d \in \mathbb{Z}^{+}$ and $r \in [0, 1]$, a \emph{context vector} is defined as a $d-$dimensional vector, of which exactly $\lfloor d r \rfloor$ elements are randomly set to $+1$, exactly $\lfloor d r \rfloor$ elements are randomly set to $-1$ and the remaining $d-2\lfloor d r \rfloor$ elements are set to $0$.   
\end{definition}

We denote the parameters $d$ and $r$ in the definition above as the \emph{dimension} and \emph{sparsity ratio} parameters respectively.

Intuitively, a context vector is defined for every \emph{atomic unit} in the corpus.  Let us denote the universe of atomic units as $U$, assumed to be a partially observed countably infinite set. In the current scenario, every unigram (a single `token') in the dataset is considered an atomic unit. Extending the definition to also include higher-order ngrams is straightforward, but was found to be unnecessary in our early empirical investigations. The universe is only partially observed because of the incompleteness (i.e. streaming, dynamic nature) of the initial corpus.

The actual vector space representation of an atomic unit is derived by defining an appropriate \emph{context} for the unit. Formally, a context is an abstract notion that is used for assigning \emph{distributional semantics} to the atomic unit. The distributional semantics hypothesis (also called \emph{Firth's axiom}) states that the semantics of an atomic unit (e.g. a word) is defined by the contexts in which it occurs \cite{distsem}.  

In this paper, we only consider \emph{short contexts} appropriate for noisy streaming data. In this vein, we define the notion of a $(u, v)$-context window below:

\begin{definition}\label{contextwindow}
Given a list $t$ of atomic units and an integer position $0<i\leq |t|$, a $(u, v)$-context window is defined by the set $S-t[i]$, where $S$ is the set of atomic units inclusively spanning positions $max(i-u, 1)$ and $min(i+v, |t|)$   
\end{definition}
\begin{figure}
\centering
\includegraphics[height=1.8in, width=2.6in]{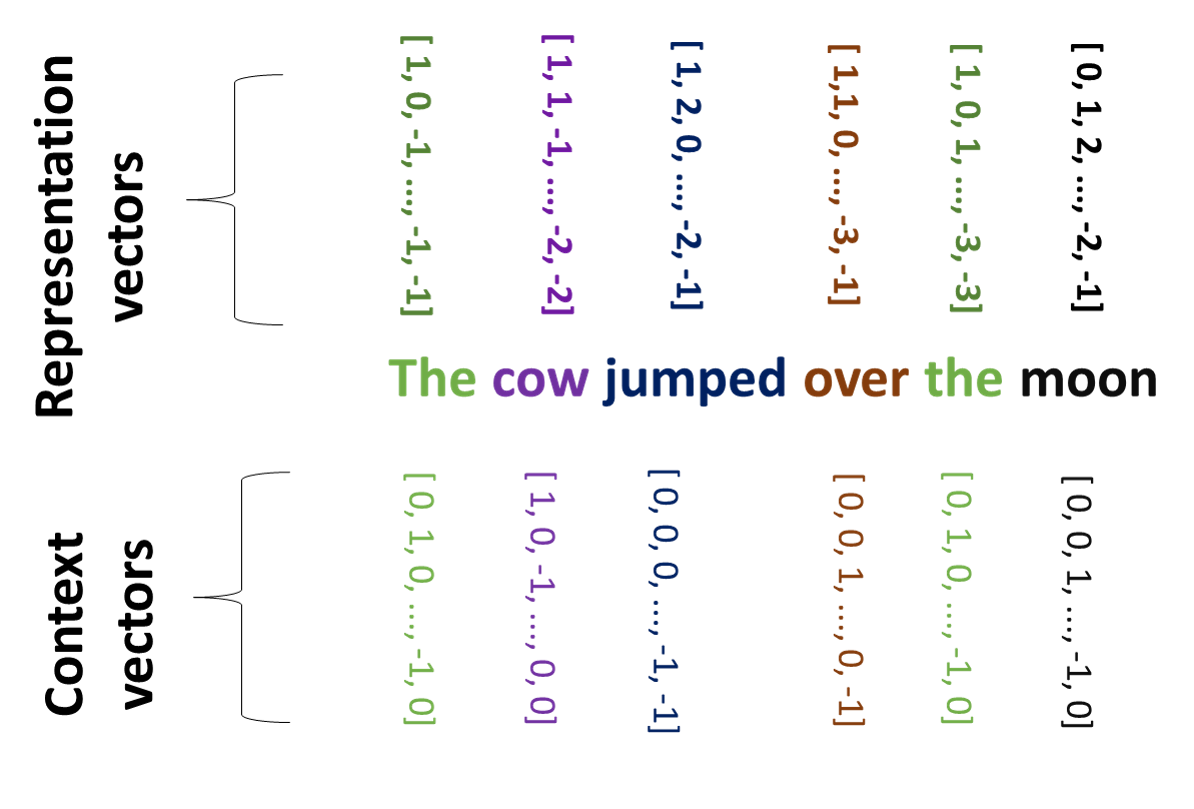}
\caption{An example illustrating the naive Random Indexing algorithm with unigram atomic units and a $(2, 2)$-context window as context}\label{riexample}
\vskip -6pt
\end{figure}
Using just these two definitions, a naive version of the RI algorithm is illustrated in Figure \ref{riexample} for the sentence `the cow jumped over the moon', assuming a $(2,2)$-context window and unigrams as atomic units. For each \emph{new} word encountered by the algorithm, a context vector (Definition \ref{contextvec}) is randomly generated, and the representation vector for the word is initialized to the 0 vector. Once generated, the context vector for the word remains fixed, but the representation vector is updated with each occurrence of the word.

The update happens as follows. Given the context of the word (ranging from a set of 2-4 words), an \emph{aggregation} is first performed on the corresponding context vectors. In Figure \ref{riexample}, for example, the aggregation is an \emph{unweighted} sum. Using the aggregated vector (denoted by the symbol $\vec{a}$), we update the representation vector using the equation below, with $\vec{w}_i$ being the representation vector derived after the $i^{th}$ occurrence of word $w$:
\begin{equation}
\vec{w}_{i+1} = \vec{w}_i+\vec{a}
\end{equation}

In principle, using this simple algorithm, we could learn a vector space representation for every atomic unit. One issue with a naive embedding of every atomic unit into a vector space is the presence of \emph{rare} atomic units. These are especially prevalent in illicit domains, not just in the form of rare words, but also as sequences of Unicode characters, sequences of HTML tags, and numeric units (e.g. phone numbers), each of which only occurs a few times (often, only once) in the corpus.

To address this issue, we define below the notion of a \emph{compound unit} that is based on a pre-specified condition.  

\begin{definition}\label{compoundunit}
Given a universe $U$ of atomic units and a binary condition $R: U \rightarrow \{True,False\}$, the \emph{compound unit} $C_R$ is defined as the largest subset of $U$ such that $R$ evaluates to True on every member of $C_R$. 
\end{definition}

\emph{Example:} For `rare' words, we could define the compound unit \emph{high-idf-units} to contain all atomic units that are below some document frequency threshold (e.g. 1\%) in the corpus. 

In our implemented prototype, we defined six \emph{mutually exclusive}\footnote{That is, an intersection of any two compound units will always be the empty set.} compound units, described and enumerated in Table \ref{table:compoundunits}. We modify the naive RI algorithm by only learning a single vector for each compound unit. Intuitively, each atomic unit $w$ in a compound unit $C$ is replaced by a special dummy symbol $w_C$; hence, after algorithm execution, each atomic unit in $C$ is represented by the single vector $\vec{w}_C$. 

\begin{table}
\centering
\caption{The compound units implemented in the current prototype}
\begin{tabular}{|p{0.95in}|p{2.1in}|} \hline
high-idf-units& Units occurring in fewer than fraction $\theta$ (by default, 1\%) of initial corpus\\ \hline
pure-num-units& Numerical units\\ \hline
alpha-num-units& Alpha-numeric units that contain at least one alphabet and one number\\ \hline
pure-punct-units& Units with only punctuation symbols\\ \hline
alpha-punct-units& Units that contain at least one alphabet and one punctuation character\\ \hline
nonascii-unicode-units& Units that only contain non-ASCII characters\\ \hline
\end{tabular}\label{table:compoundunits}
\end{table}

\subsection{Applying High-Recall Recognizers}\label{recognizers}
 
For a given attribute (e.g. \emph{City}) and a given corpus, we define a recognizer as a function that, if known, can be used to exactly determine the instances of the attribute occurring in the corpus. Formally,
\begin{definition}
A recognizer $R_A$ for attribute $A$ is a function that takes a list $t$ of tokens and positions $i$ and $j >= i$ as inputs, and returns \emph{True} if the tokens contiguously spanning $t[i]:t[j]$ are instances of $A$, and \emph{False} otherwise.  
\end{definition}
It is important to note that, per the definition above, a recognizer cannot annotate \emph{latent} instances that are not directly observed in the list of tokens.  

 Since the `ideal' recognizer is not known, the broad goal of IE is to devise models that approximate it (for a given attribute) with high accuracy. Accuracy is typically measured in terms of precision and recall metrics. We formulate a two-pronged approach whereby, rather than develop a single recognizer that has both high precision and recall (and requires considerable expertise to design), we first obtain a list of candidate annotations that have high recall in expectation, and then use supervised classification in a second step to improve precision of the candidate annotations. 
 
More formally, let $R_A$ be denoted as an $\eta$-recall recognizer if the expected recall of $R_A$ is at least $\eta$. Due to the explosive growth in data, many resources on the Web can be used for bootstrapping recognizers that are `high-recall' in that $\eta$ is in the range of 90-100\%. The high-recall recognizers currently used in the prototype described in this paper (detailed further in Section \ref{system}) rely on knowledge bases (e.g. GeoNames) from Linked Open Data \cite{lod}, dictionaries from the Web and broad heuristics, such as regular expression extractors, found in public Github repositories. In our experience, we found that even students with basic knowledge of GitHub and Linked Open Data sources are able to construct such recognizers. One important reason why constructing such recognizers is relatively hassle-free is because they are typically \emph{monotonic} i.e. new heuristics and annotation sources can be freely integrated, since we do not worry about precision at this step.

We note that in some cases, domain knowledge alone is enough to guarantee 100\% recall for well-designed recognizers for certain attributes. In HT, this is true for location attributes like city and state, since advertisements tend to state locations without obfuscation, and we use GeoNames, an exhaustive knowledge base of locations, as our recognizer.
Manual inspection of the ground-truth data showed that the recall of utilized recognizers for attributes like \emph{Name} and \emph{Age} are also high (in many cases, 100\%). Thus, although 100\% recall cannot be \emph{guaranteed} for any recognizer, it is still reasonable to assume that $\eta$ is high.

A much more difficult problem is engineering a recognizer to simultaneously achieve high recall \emph{and} high precision. Even for recognizers based on curated knowledge bases like GeoNames, many non-locations get annotated as locations. For example, the word `nice' is a city in France, but is also a commonly occurring adjective. Other common words like `for', `hot', `com', `kim' and `bella' also occur in GeoNames as cities and would be annotated. Using a standard Named Entity Recognition system does not always work because of the language modeling problem (e.g. missing capitalization) in illicit domains. In the next section, we show how the context surrounding the annotated word can be used to classify the annotation as correct or incorrect. We note that, because the recognizers are high-recall, a successful classifier would yield both high precision and recall.

\subsection{Supervised Contextual Classifier}\label{validate}
\begin{figure}
\centering
\includegraphics[height=1.4in, width=3.3in]{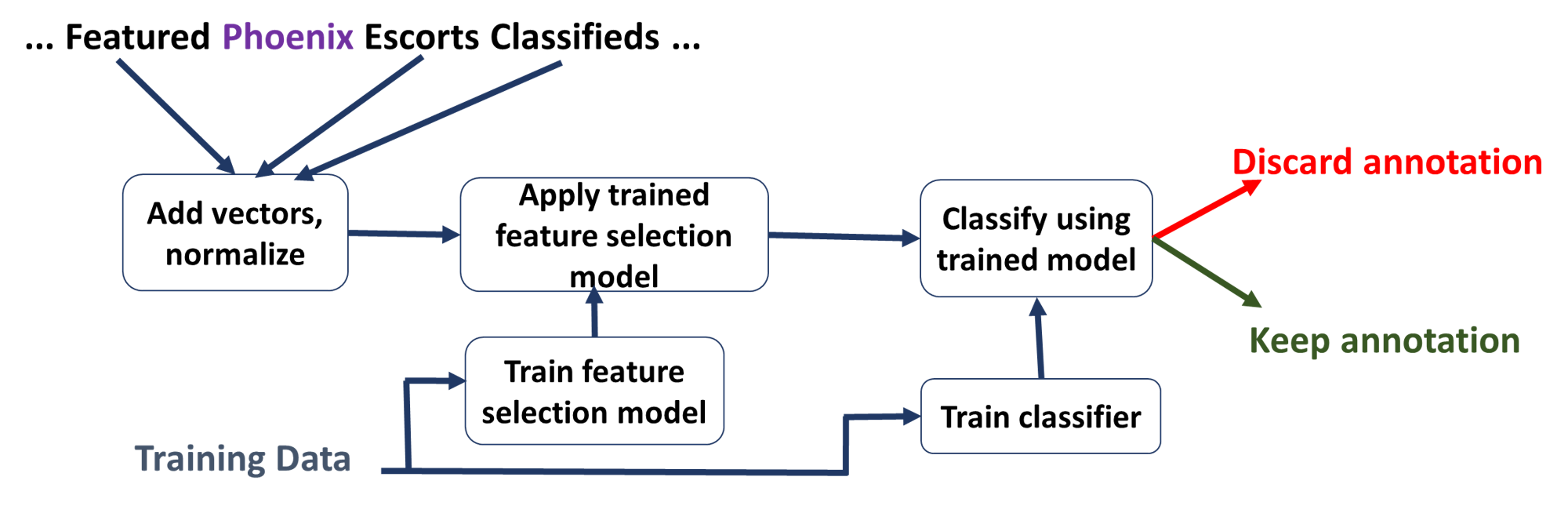}
\caption{An illustration of supervised contextual classification on an example annotation (`Phoenix')}\label{contextualclassifier}
\vskip -6pt
\end{figure}

To address the precision problem, we train a classifier using contextual features. Rather than rely on a domain expert to provide a set of hand-crafted features, we derive a feature vector per candidate annotation using the notion of a context window (Definition \ref{contextwindow}) and the word representation vectors derived in Section \ref{embeddings}. This process of \emph{supervised contextual classification} is illustrated in Figure \ref{contextualclassifier}. 

Specifically, for each annotation (which could comprise \emph{multiple} contiguous tokens e.g. `Salt Lake City' in the list of tokens representing the website) annotated by a recognizer, we consider the tokens in the $(u, v)$-context window around the annotation. We aggregate the vectors of those tokens into a single vector by performing an unweighted sum, followed by $l2$-normalization. We use this aggregate vector as the \emph{contextual feature vector} for that annotation. Note that, unlike the representation learning phase, where the surrounding \emph{context vectors} were aggregated into an existing representation vector, the contextual feature vector is obtained by summing the actual representation vectors. 

For \emph{each} attribute, a supervised machine learning classifier (e.g. random forest) is trained using between 12-120 labeled annotations, and for new data, the remaining annotations can be classified using the trained classifier. Although the number of dimensions in the feature vectors is quite low compared to \emph{tf-idf} vectors (hundreds vs. millions), a second round of dimensionality reduction can be applied by using (either supervised or unsupervised) feature selection for further empirical benefits (Section \ref{evaluations}).

\section{Evaluations}\label{evaluations}

\subsection{Datasets and Ground-truths}
We train the word representations on four real-world human trafficking datasets of increasing size, the details of which are provided in Table \ref{embeddingdatasets}.  Since we assume a `streaming' setting in this paper, each larger dataset in Table \ref{embeddingdatasets} is a strict superset of the smaller datasets. The largest dataset is itself a subset of the overall human trafficking corpus that was scraped as part of research conducted in the DARPA MEMEX program\footnote{\url{http://www.darpa.mil/program/memex}}.

\begin{table}
\centering
\caption{Four human trafficking corpora for which word representations are (independently) learned}
\begin{tabular}{|p{0.5in}|p{0.9in}|p{0.7in}|p{0.7in}|} \hline
Name&Num. websites&Total word count& Unique word count\\ \hline
D-10K & 10,000& 2,351,036&1,030,469\\ \hline
D-50K & 50,000& 11,758,647 &5,141,375\\ \hline
D-100K & 100,000& 23,536,935&10,277,732\\ \hline
D-ALL & 184,132& 43,342,278&18,940,260\\ \hline
\end{tabular}\label{embeddingdatasets}
\end{table}

Since ground-truth extractions for the corpus are unknown, we randomly sampled websites from the overall corpus\footnote{Hence, it is possible that there are websites in the ground-truth that are not part of the corpora in Table \ref{embeddingdatasets}.}, applied four high-recall recognizers described in Section \ref{system}, and for each annotated set, manually verified whether the extractions were correct or incorrect for the corresponding attribute. The details of this sampled ground-truth are captured in Table \ref{groundtruthdatasets}. Each annotation set is named using the format \emph{GT-\{RawField\}-\{AnnotationAttribute\}}, where \emph{RawField} can be either the HTML title or the scraped text (Section \ref{preprocessing}). and \emph{AnnotationAttribute} is the attribute of interest for annotation purposes.
\begin{table}
\centering
\caption{Five ground-truth datasets on which the classifier (Section \ref{validate}) and baselines are evaluated}
\begin{tabular}{|p{1.0in}|p{0.3in}|p{0.3in}|p{1.1in}|} \hline
Name&Pos. ann.&Neg. ann.& Recognizer Used\\ \hline

GT-Text-City & 353 & 15,783 &GeoNames-Cities\\ \hline
GT-Text-State & 100& 16,036&GeoNames-States\\ \hline
GT-Title-City & 37 & 513 &GeoNames-Cities\\ \hline
GT-Text-Name & 162& 14,337&Dictionary-Names\\ \hline
GT-Text-Age & 116& 14,306&RegEx-Ages\\ \hline
\end{tabular}\label{groundtruthdatasets}
\end{table}

\vfill\eject
\subsection{System}\label{system}
The overall system requires developing two components for each attribute: a high-recall recognizer and a classifier for pruning annotations. We developed four high-recall recognizers, namely \emph{GeoNames-Cities, GeoNames-States, RegEx-Ages} and \emph{Dictionary-Names}. The first two of these relies on the freely available GeoNames\footnote{\url{http://www.geonames.org/}} dataset \cite{geonames}; we use the entire dataset for our experiments, which involves modeling each GeoNames dictionary as a trie, owing to its large memory footprint. For extracting ages, we rely on simple regular expressions and heuristics that were empirically verified to capture a broad set of age representations\footnote{The age extractors we used are also available in the Github repository accessed at \url{https://github.com/usc-isi-i2/dig-age-extractor}}. For the name attribute, we gather freely available \emph{Name} dictionaries on the Web, in multiple countries and languages, and use the dictionaries\footnote{For replication, the full set of dictionaries used may be accessed at \url{https://github.com/usc-isi-i2/dig-dictionaries/tree/master/person-names}} in a case-insensitive recognition algorithm to locate names in the raw field (i.e. text or title). 

\subsection{Baselines}
\begin{table}
\centering
\caption{Stanford NER features that were used for re-training the model on our annotation sets}
\begin{tabular}{|p{1.3in}|p{1.4in}|} \hline
useClassFeature=true& useNext=true\\ \hline
useWord=true& useSequences=true\\ \hline
useNGrams=true& usePrevSequences=true\\ \hline
noMidNGrams=true& maxLeft=1\\ \hline
useDisjunctive=true&useTypeSeqs=true\\ \hline
maxNGramLeng=6&useTypeSeqs2=true\\ \hline
usePrev=true&useTypeySequences=true\\ \hline
wordShape=chris2useLC & \\ \hline
\end{tabular}\label{nerfeatures}
\end{table}
We use different variants of the Stanford Named Entity Recognition system (NER) as our baselines \cite{stanfordner}. For the first set of baselines, we use two pre-trained models trained on different English language corpora\footnote{Details are available at \url{http://nlp.stanford.edu/software/CRF-NER.shtml#Models}}. Specifically, we use the 3-Class and 4-Class pre-trained models\footnote{In all Stanford NER pre-trained models, the \emph{distributional similarity} option was enabled, which is known to boost F1-Measure scores.}. We use the LOCATION class label for determining city and state annotations, and the PERSON label for name annotations. Unfortunately, there is no specific label corresponding to age annotations in the pre-trained models; hence, we do not use the pre-trained models as age annotation baselines.

It is also possible to \emph{re-train} the underlying NER system on a new dataset. For the second set of baselines, therefore, we re-train the NER models by randomly sampling 30\% and 70\% of each annotation set in Table \ref{groundtruthdatasets} respectively, with the remaining annotations used for testing. The features and values that were employed in the re-trained models are enumerated in Table \ref{nerfeatures}. Further documentation on these feature settings may be found on the \emph{NERFeatureFactory} page\footnote{Documentation accessed at \url{http://nlp.stanford.edu/nlp/javadoc/javanlp/edu/stanford/nlp/ie/NERFeatureFactory.html}}. All training and testing experiments were done in ten independent trials\footnote{When evaluating the pre-trained models, the training set is ignored and only the testing set is classified.}. We use default parameter settings, and report average results for each experimental run. Experimentation using other configurations, features and values is left for future studies. 

\subsection{Setup and Parameters}\label{setup}
{\bf Parameter tuning} System parameters were set as follows. The number of dimensions in Definition \ref{contextvec} was set at 200, and the sparsity ratio was set at 0.01. These parameters are similar to those suggested in previous word representation papers; they were also found to yield intuitive results on semantic similarity experiments (described further in Section \ref{discussion}). To avoid the problem of rare words, numbers, punctuation and tags, we used the six compound unit classes earlier described in Table \ref{table:compoundunits}. In all experiments where defining a context was required, we used symmetric $(2,2)$-context windows; using bigger windows was not found to offer much benefit. We trained a random forest model with default hyperparameters (10 trees, with Gini Impurity as the split criterion) as the supervised classifier, used supervised k-best feature selection with $k$ set to 20 (Section \ref{validate}), and with the Analysis of Variance (ANOVA) F-statistic between class label and feature used as the feature scoring function.

Because of the \emph{class skew} in Table \ref{groundtruthdatasets} (i.e. the `positive' class is typically much smaller than the `negative' class) we oversampled the positive class for balanced training of the supervised contextual classifier.   

{\bf Metrics} The metrics used for evaluating IE effectiveness are Precision, Recall and F1-measure. 

{\bf Implementation} In the interests of demonstrating a reasonably lightweight system, all experiments in this paper were run on a serial iMac with a 4 GHz Intel core i7 processor and 32 GB RAM. All code (except the Stanford NER code) was written in the Python programming language, and has been made available on a public Github repository\footnote{\url{https://github.com/mayankkejriwal/fast-word-embeddings}} with documentation and examples. We used Python's Scikit-learn library (v0.18) for the machine learning components of the prototype.

\subsection{Results}
\begin{table*}
\centering
\caption{Comparative results of three systems on precision (P), recall (R) and F1-Measure (F) when training percentage is 30. For the pre-trained baselines, we only report the best results across all applicable models}
\begin{tabular}{|p{1.4in}|p{1.6in}|p{1.7in}|p{1.7in}|} \hline
Ground-truth Dataset&Our System (P/R/F) &Re-trained Baseline (P/R/F)&Pre-trained Baseline (P/R/F)\\ \hline
GT-Text-City & 0.5207/0.5050/0.5116
 & {\bf 0.9855}/0.1965/0.3225
 &0.7206/{\bf 0.7406}/{\bf 0.7299}
\\ \hline
GT-Text-State & {\bf 0.7852}/0.6887/{\bf 0.7310}
& 0.64/0.0598/0.1032
&0.2602/{\bf 0.8831}/0.3993
\\ \hline
GT-Title-City &  0.5374/0.5524/0.5406
 & {\bf 0.8633}/0.1651/0.2685 &0.8524/{\bf 0.7341}/{\bf 0.7852}
\\ \hline
GT-Text-Name &  0.7201/{\bf 0.5850}/{\bf 0.6388}
& {\bf 1}/0.2103/0.3351
&0/0/0\\ \hline
GT-Text-Age & 0.8993/{\bf 0.9156}/{\bf 0.9068}
& {\bf 0.9102}/0.7859/0.8412
&N/A
\\ \hline \hline
\emph{Average} & 0.6925/{\bf 0.6493}/{\bf 0.6658}
& {\bf 0.8798}/0.2835/0.3741
&0.4583/0.5895/0.4786
\\ \hline
\end{tabular}\label{results30}
\end{table*}

\begin{table*}
\centering
\caption{Comparative results of three systems when training percentage is 70}
\begin{tabular}{|p{1.4in}|p{1.6in}|p{1.7in}|p{1.7in}|} \hline
Ground-truth Dataset&Our System (P/R/F) &Re-trained Baseline (P/R/F)&Pre-trained Baseline (P/R/F)\\ \hline
GT-Text-City & 0.5633/0.6081/0.5841
 & {\bf 0.9434}/0.3637/0.5000
 &0.6893/{\bf 0.7401}/{\bf 0.7128}
\\ \hline
GT-Text-State & {\bf 0.7916}/0.7269/{\bf 0.7502}
& 0.7833/0.2128/0.2971
&0.1661/{\bf 0.7830}/0.2655
\\ \hline
GT-Title-City & 0.6403/{\bf 0.6667}/0.6437
 & {\bf 0.9417}/0.3333/0.4790
 &0.9133/0.6384/{\bf 0.7289}
\\ \hline
GT-Text-Name & 0.7174/{\bf 0.6818}/{\bf 0.6960}
& {\bf 1}/0.3747/0.5140
&0/0/0\\ \hline
GT-Text-Age & 0.9252/{\bf 0.9273}/{\bf 0.9251}
& {\bf 0.9254}/0.8454/0.8804
&N/A \\ \hline \hline
\emph{Average} & 0.7276/{\bf 0.7222}/{\bf 0.7198}
& {\bf 0.9188}/0.4260/0.5341
&0.4422/0.5404/0.4268
\\ \hline
\end{tabular}\label{results70}
\end{table*}
{\bf Performance against baselines} Table \ref{results30} illustrates system performance on Precision, Recall and F1-Measure metrics against the re-trained and pre-trained baseline models, where the re-trained model and our approach were trained on 30\% of the annotations in Table \ref{groundtruthdatasets}. We used the word representations derived from the D-ALL corpus. On average, the proposed system performs the best on F1-Measure and recall metrics. The re-trained NER is the most precise system, but at the cost of much less recall ($<$30\%). The good performance of the pre-trained baseline on the \emph{City} attribute demonstrates the importance of having a large training corpus, even if the corpus is not directly from the test domain. On the other hand, the complete failure of the pre-trained baseline on the \emph{Name} attribute illustrates the dangers of using out-of-domain training data. As noted earlier, language models in illicit domains can significantly differ from natural language models; in fact, names in human trafficking websites are often represented in a variety of misleading ways.

Recognizing that 30\% training data may constitute a sample size too small to make reliable judgments, we also tabulate the results in Table \ref{results70} when the training percentage is set at 70. Performance improves for both the re-trained baseline and our system. Performance declines for the pre-trained baseline, but this may be because of the sparseness of positive annotations in the smaller test set. 

We also note that performance is relatively well-balanced for our system; on all datasets and all metrics, the system achieves scores greater than 50\%. This suggests that our approach has a degree of robustness that the CRFs are unable to achieve; we believe that this is a direct consequence of using contextual word representation-based feature vectors.   

\begin{figure}
\centering
\includegraphics[height=1.8in, width=2.7in]{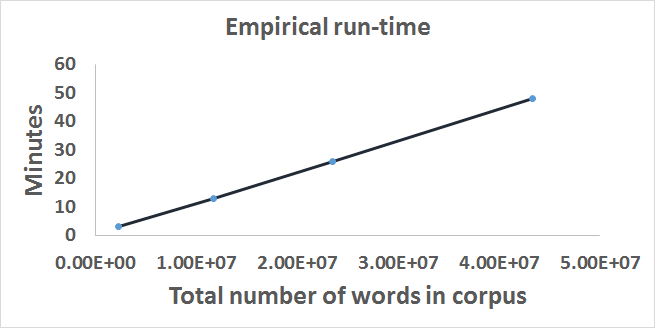}
\caption{Empirical run-time of the adapted random indexing algorithm on the corpora in Table \ref{embeddingdatasets}}\label{fig:runtime}
\vskip -6pt
\end{figure}

{\bf Runtimes} We recorded the runtimes for learning word representations using the random indexing algorithm described earlier on the four datasets in Table \ref{embeddingdatasets}, and plot the runtimes in Figure \ref{fig:runtime} as a function of the total number of words in each corpus. 
In agreement with the expected theoretical time-complexity of random indexing, the empirical run-time is linear in the number of words, for fixed parameter settings. More importantly, the absolute times show that  the algorithm is extremely lightweight: on the D-ALL corpus, we are able to learn representations in under an hour. 

We note that these results do not employ any obvious parallelization or the multi-core capabilities of the machine. The linear scaling properties of the algorithm show that it can be used even for very large Web corpora. In future, we will investigate an implementation of the algorithm in a distributed setting.

{\bf Robustness to corpus size and quality} One issue with using large corpora to derive word representations is \emph{concept drift}. The D-ALL corpora, for example, contains tens of different Web domains, even though they all pertain to human trafficking. An interesting empirical issue is whether a smaller corpus (e.g. D-10K or D-50K) contains enough data for the derived word representations to converge to reasonable values. Not only would this alleviate initial training times, but it would also partially compensate for concept drift, since it would be expected to contain fewer unique Web domains. 

Tables \ref{initcorpus1} and \ref{initcorpus2} show that such generalization is possible. The best F1-Measure performance, in fact, is achieved for D-10K, although the average F1-Measures vary by a margin of less than 2\% on all cases. We cite this as further evidence of the robustness of the overall approach.
\begin{table}
\centering
\caption{A comparison of F1-Measure scores of our system (30\% training data), with word representations trained on different corpora}
\begin{tabular}{|p{0.9in}|p{0.37in}|p{0.37in}|p{0.45in}|p{0.4in}|} \hline
Ground-truth&D-10K&D-50K&D-100K&D-ALL\\ \hline

GT-Text-City &0.4980 &0.5058 & 0.4909&{\bf 0.5116} \\ \hline
GT-Text-State & 0.7362& 0.7385& {\bf 0.7526}&0.7310\\ \hline
GT-Title-City & {\bf 0.6148}& 0.5638& 0.5061&0.5406\\ \hline
GT-Text-Name & 0.6756& 0.6881& {\bf 0.6920}&0.6388\\ \hline
GT-Text-Age & {\bf 0.9387}& 0.9364& 0.9171&0.9068\\ \hline \hline
\emph{Average} & {\bf 0.6927} & 0.6865 & 0.6717 & 0.6658 \\ \hline
\end{tabular}\label{initcorpus1}
\end{table}

\begin{figure}
\centering
\includegraphics[height=1.6in, width=3.1in]{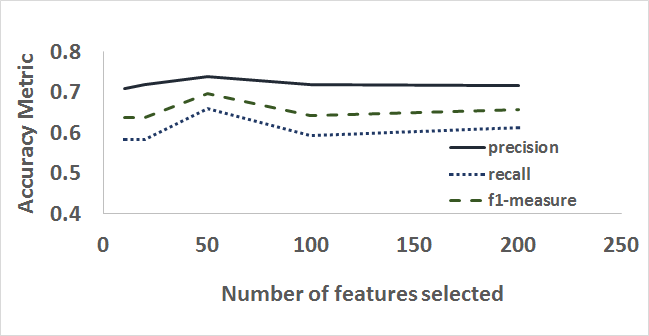}
\caption{Effects of additional feature selection on the \emph{GT-Text-Name} dataset (30\% training data)}\label{fs1}
\vskip -6pt
\end{figure}

{\bf Effects of feature selection} Finally, we evaluate the effects of feature selection in Figure \ref{fs1} on the \emph{GT-Text-Name} dataset, with training percentage set\footnote{Results on the other datasets were qualitatively similar; we omit full reproductions herein.} at 30. The results show that, although performance is reasonably stable for a wide range of $k$, some feature selection is necessary for better generalization.

\begin{table}
\centering
\caption{A comparison of F1-Measure scores of our system (70\% training data), with word representations trained on different corpora}
\begin{tabular}{|p{0.9in}|p{0.37in}|p{0.37in}|p{0.45in}|p{0.4in}|} \hline
Ground-truth&D-10K&D-50K&D-100K&D-ALL\\ \hline

GT-Text-City &{\bf 0.5925} & 0.5781& 0.5716&0.5841 \\ \hline
GT-Text-State & 0.7357& {\bf 0.7641}& 0.7246 &0.7502\\ \hline
GT-Title-City & 0.6424& 0.6428& 0.6364&{\bf 0.6437}\\ \hline
GT-Text-Name & {\bf 0.7665}& 0.7091& 0.7333&0.6960\\ \hline
GT-Text-Age & 0.9311&{\bf 0.9634} & 0.9347&0.9251\\ \hline \hline
\emph{Average} & {\bf 0.7336} & 0.7315 & 0.7201 & 0.7198 \\ \hline
\end{tabular}\label{initcorpus2}
\end{table}

\subsection{Discussion}\label{discussion}
Table \ref{examples} contains some examples (in bold) of cities that got correctly extracted, with the bold term being assigned the highest score by the contextual classifier that was trained for cities. The examples provide good evidence for the kinds of variation (i.e. concept drift) that are often observed in real-world human trafficking data over multiple Web domains. Some domains, for example, were found to have the same kind of structured format as the second row of Table \ref{examples} (i.e. \emph{Location:} followed by the actual locations), but many other domains were far more heterogeneous. 
\begin{figure}
\centering
\includegraphics[height=2.1in, width=3.3in]{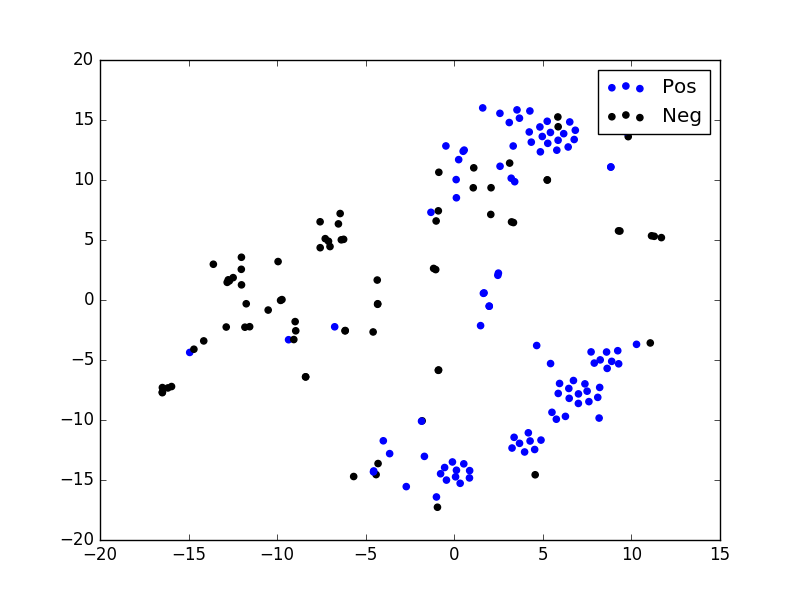}
\caption{Visualizing city contextual classifier inputs (with colors indicating ground-truth labels) using the t-SNE tool}\label{fig:tsne}
\vskip -6pt
\end{figure}
\begin{table}
\centering
\caption{Some representative examples of correct city extractions using the proposed method}
\begin{tabular}{|p{3.0in}|} \hline
\ldots 1332 SOUTH 119TH STREET, {\bf OMAHA} NE 68144 \ldots \\ \hline
\ldots Location: Bossier City/{\bf Shreveport} \ldots \\ \hline
\ldots to service the areas of {\bf Salt Lake City} Ogden,Farmington,Centerville,Bountiful \ldots \\ \hline
\ldots 4th August 2015 in {\bf rochester} ny, new york \ldots \\ \hline
\ldots willing to Travel ( Cali, {\bf Miami}, New York, Memphis \ldots \\ \hline
\ldots More girls from {\bf Salt Lake City}, UT \ldots \\ \hline
\end{tabular}\label{examples}
\end{table}

The results in this section also illustrate the merits of unsupervised feature engineering and contextual supervision. In principle, there is no reason why the word representation learning module in Figure \ref{fig:approach} cannot be replaced by a more adaptive algorithm like Word2vec \cite{word2vec}. We note again that, before applying such algorithms, it is important to deal with the heterogeneity problem that arises from having many different Web domains present in the corpus. While
earlier results in this section (Tables \ref{initcorpus1} and \ref{initcorpus2}) showed that random indexing is reasonably stable as more websites are added to the corpus, we also verify this robustness \emph{qualitatively} using a few domain-specific examples in Table \ref{embeddingExamples}. We ran the qualitative experiment as follows: for each seed token (e.g. `tall'), we searched for the two nearest neighbors in the semantic space induced by random indexing by applying cosine similarity, using two different word representation datasets (D-10K and D-ALL). As the results in Table \ref{embeddingExamples} show, the induced distributional semantics are stable; even when the nearest neighbors are different (e.g. for `tall'), their semantics still tend to be similar.
\begin{table}
\centering
\caption{Examples of semantic similarity using random indexing vectors from D-10K and D-ALL}
\begin{tabular}{|p{0.8in}|p{0.95in}||p{1.1in}|} \hline
{\bf Seed-token} & {\bf D-10K} & {\bf D-ALL} \\ \hline
tall & figure, attractive & fit, cute \\ \hline
florida & california, ohio & california, texas \\ \hline
green & blue, brown & blue, brown \\ \hline
attractive & fit, figure & elegant, fit \\ \hline
open-minded & playful, sweet & passionate, playful \\ \hline
\end{tabular}\label{embeddingExamples}
\end{table}

Another important point implied by both the qualitative and quantitative results on D-10K is that random indexing is able to generalize quickly even on \emph{small} amounts of data. 
To the best of our knowledge, it is currently an open question (theoretically and empirically), at the time of writing, whether state-of-the-art \emph{neural} embedding-based word representation learners can (1) generalize on small quantities of data, especially in a single epoch (`streaming data') (2) adequately compensate for concept drift with the same degree of robustness, and in the same lightweight manner, as the random indexing method that we adapted and evaluated in this paper. A broader empirical study on this issue is warranted. 

Concerning contextual supervision, we qualitatively visualize the inputs to the contextual city classifier using the t-SNE tool \cite{tsne}. We use the ground-truth labels to determine the color of each point in the projected 2d space. The plot in Figure \ref{fig:tsne} shows that there is a reasonable separation of labels; interestingly there are also `sub-clusters' among the positively labeled points. Each sub-cluster provides evidence for a similar context; the number of sub-clusters even in this small sample of points again illustrates the heterogeneity in the underlying data.  

A last issue that we mention is the generalization of the method to more unconventional attributes than the ones evaluated herein. In ongoing work, we have experimented with more domain-specific attributes such as \emph{ethnicity} (of escorts), and have achieved similar performance. In general, the presented method is applicable whenever the context around the extraction is a suitable clue for disambiguation. 


\section{Conclusion}\label{conclusion}
 In this paper, we presented a lightweight, feature-agnostic Information Extraction approach that is suitable for illicit Web domains. Our approach relies on unsupervised derivation of word representations from an initial corpus, and the training of a supervised contextual classifier using external high-recall recognizers and a handful of manually verified annotations. Experimental evaluations show that our approach can outperform feature-centric CRF-based approaches for a range of generic attributes. Key modules of our prototype are publicly available (see footnote 15) and can be efficiently bootstrapped in a serial computing environment. Some of these modules are already being used in real-world settings. For example, they were recently released as tools for graduate-level participants in the End Human Trafficking hackathon organized by the office of the District Attorney of New York\footnote{\url{https://ehthackathon.splashthat.com/}}. At the time of writing, the system is being actively maintained and updated. 
 
 {\bf Acknowledgements} The authors gratefully acknowledge the efforts of Lingzhe Teng, Rahul Kapoor and Vinay Rao Dandin, for sampling and producing the ground-truths in Table \ref{groundtruthdatasets}. This research is supported by the Defense Advanced Research Projects
Agency (DARPA) and the Air Force Research Laboratory (AFRL) under contract number FA8750-
14-C-0240. The views and conclusions contained herein are those of the authors and should not
be interpreted as necessarily representing the official policies or endorsements, either expressed
or implied, of DARPA, AFRL, or the U.S. Government.

%
\bibliographystyle{abbrv}
\bibliography{sigproc}  
%
%

\end{document}